\documentclass[10pt,twocolumn,letterpaper]{article}

\usepackage{cvpr}              
\definecolor{cvprblue}{rgb}{0.21,0.49,0.74}
\usepackage[pagebackref,breaklinks,colorlinks,allcolors=cvprblue]{hyperref}
\usepackage{array}

\title{DREAM: Dual-Memory Reasoning for Anticipatory 3D Perception in Autonomous Driving}

\author{Sriram Mandalika\\
Hasso Plattner Institute\\
Potsdam, Germany\\
{\tt\small sriram.mandalika@student.hpi.uni-potsdam.de}
\and
Athira Nambiar\thanks{Corresponding author.}\\
SRM Institute of Science and Technology\\
Kattankulathur, India\\
{\tt\small athiram@srmist.edu.in}
}

\begin{document}
\maketitle
\begin{abstract}
Autonomous driving systems must not only interpret the current scene but also anticipate perceptual challenges such as occlusions, sparse observations, and recurring failure patterns. However, most existing LiDAR semantic segmentation methods operate in a frame-centric manner, processing each scan independently and discarding past contextual experience, which limits robustness under dynamic scene changes. We propose \textbf{DREAM}, a \textbf{D}ual-memory \textbf{REA}soning framework for continual LiDAR se\textbf{M}antic segmentation that reframes perception as an experience-driven and anticipatory process. DREAM maintains two complementary memory banks in a strictly online setting: a latent memory that stores compact semantic abstractions of previously observed scenes, and an error memory that records representations associated with uncertain predictions. At each timestep, relevant memory entries are retrieved via cosine similarity and integrated into the feature space through a lightweight modulation mechanism, enabling the model to reinforce consistent semantic patterns while suppressing recurring failure modes. The backbone remains frozen and no past scans are replayed, ensuring computational efficiency and bounded memory growth. Extensive experiments on multiple large-scale LiDAR benchmarks demonstrate that DREAM achieves state-of-the-art performance, with consistent improvements on dynamic and small-scale objects, highlighting the effectiveness of persistent and error-aware memory for robust long-horizon perception.
\end{abstract}    
\section{Introduction}
\label{sec:intro}

Semantic scene understanding is a core capability for autonomous systems operating in complex environments such as autonomous driving. Among perception tasks, semantic segmentation provides dense structured representations of the surrounding scene that are critical for downstream planning and decision-making. However, reliable autonomy requires perception systems that remain robust under challenging sensing conditions such as occlusions, sparse measurements, and dynamic scene changes. Despite strong progress in deep learning–based perception, most LiDAR semantic segmentation methods remain fundamentally reactive, processing each scan independently or within short temporal windows \cite{Dewan2019DeepTemporalSegTC}. This frame-centric formulation prevents perception systems from leveraging past experience to anticipate recurring perceptual challenges.

In real-world deployment, LiDAR observations arrive as a continuous stream where each scan provides only a partial and noisy view of the environment. Measurements are often affected by occlusions, sensor noise, and long-range sparsity. Under such conditions, treating frames independently \cite{Cheng2021AF2S3NetAF, Milioto2019RangeNetF, Tang2020SearchingE3, Xu2020SqueezeSegV3SC, Yan20222DPASS2P, Zhu2021CylindricalAA} leads to brittle predictions and temporal inconsistencies, particularly for distant or rarely observed structures. In contrast, human perception relies on accumulated experience: past observations inform future interpretation and help anticipate recurring patterns of uncertainty \cite{MODIRSHANECHI2023102758}. Enabling perception models to accumulate and exploit such experience remains a key challenge for robust long-horizon scene understanding.

Three-dimensional representations provide a natural foundation for storing perceptual experience. Unlike image projections, 3D space preserves metric structure, allowing memory to maintain consistent object scales and spatial relationships across viewpoints. This enables stable semantic priors, more uniform representation across near and far regions, and improved reasoning about occlusions and motion. Regions that become temporarily unobserved due to occlusions can therefore be retained in memory without interference from newly observed surfaces. However, maintaining spatial memory alone does not fully address a key limitation of deployed perception systems: the tendency to make similar errors under recurring sensing conditions repeatedly.

Recent work has begun exploring memory-augmented perception for LiDAR segmentation. Earlier approaches, such as Cylinder3D~\cite{Zhu2021CylindricalAA} and Meta-RangeSeg~\cite{Wang2022MetaRangeSegLS}, primarily focus on improving feature representations and spatial-temporal aggregation for semantic segmentation, but they do not explicitly maintain persistent memory across observations. MemorySeg~\cite{Li2023MemorySegOL}, for example, maintains a persistent voxel memory that aggregates information across streaming LiDAR scans and improves understanding of long-range and occluded regions. While effective, such approaches primarily accumulate spatial context and treat memory as a passive repository of past observations. In practice, perception systems frequently encounter recurring failure patterns, such as systematic uncertainty under severe sparsity or occlusion. Conventional architectures lack mechanisms to explicitly represent these failure modes and use them to guide future predictions.

To address this limitation, we propose \textbf{DREAM}, a dual-memory framework for continual LiDAR semantic segmentation that augments perception with both perceptual and error-aware memory. DREAM maintains two compact latent memory banks operating in a strictly online setting. A \emph{latent perceptual memory} stores compressed semantic descriptors of previously observed scenes, while an \emph{error memory} stores latent representations associated with uncertain predictions. At each timestep, a global latent descriptor extracted from the current scan retrieves relevant entries from both memory banks using cosine-similarity addressing. The retrieved latent and error memories are projected into the dense feature space and integrated through a lightweight modulation mechanism prior to classification. This allows the network to reinforce reliable semantic patterns while suppressing conditions that previously produced low-confidence predictions.

The proposed framework DREAM, processes one scan at a time, maintains bounded memory capacity, and avoids computationally expensive attention mechanisms or long-horizon backpropagation. The backbone encoder remains frozen, while lightweight memory modulation layers condition perception on retrieved experience. This design enables temporally aware semantic segmentation while maintaining efficiency suitable for real-time autonomous driving scenarios. The effectiveness of the proposed DREAM architecture is substantiated through comprehensive evaluations on three benchmark datasets: SemanticKITTI\cite{behley2019iccv}, nuScenes\cite{Caesar2019nuScenesAM}, and PandaSet\cite{Xiao2021PandaSetAS}.

In summary, the key contributions of this work are as follows:
\begin{itemize}
    \item Proposal of  DREAM, a Dual-Memory Reasoning for Anticipatory 3D Perception for Autonomous Driving.

    \item We introduce an error-aware memory mechanism that explicitly models recurring uncertainty patterns and helps mitigate future perceptual failures
    
    \item Proposal of a Dual Memory Modulation scheme for continual LiDAR semantic segmentation that integrates perceptual and error-aware memories to enable experience-conditioned perception, one of its first kinds.
    
\end{itemize}
\section{Related Works}
\label{sec:RW}

In this section, we review existing LiDAR semantic segmentation methods and discuss approaches that incorporate temporal reasoning for streaming LiDAR perception.

\subsection{LiDAR Semantic Segmentation}
LiDAR semantic segmentation methods can be broadly categorized based on the underlying data representation: point-based~\cite{Qi2016PointNetDL,Qi2017PointNetDH,Hu2019RandLANetES,Thomas2019KPConvFA}, projection-based~\cite{Xu2020SqueezeSegV3SC,Cortinhal2020SalsaNextFS,Zhang2020PolarNetAI}, voxel-based~\cite{Cheng2021AF2S3NetAF,Zhu2021CylindricalAA,Hou2022PointtoVoxelKD}, and hybrid approaches that combine multiple representations~\cite{Tang2020SearchingE3,Xu2021RPVNetAD}.

Point-based methods~\cite{Qi2016PointNetDL,Qi2017PointNetDH,Hu2019RandLANetES,Thomas2019KPConvFA} operate directly on raw point clouds and learn point-wise features using permutation-invariant networks. While effective, many of these approaches rely on aggressive downsampling to maintain computational efficiency. For example, KPConv~\cite{Thomas2019KPConvFA} introduces deformable point convolutions that significantly improve segmentation performance, but at the cost of increased computational complexity.

Projection-based methods~\cite{Xu2020SqueezeSegV3SC,Cortinhal2020SalsaNextFS,Zhang2020PolarNetAI} project 3D point clouds into structured 2D representations such as range-view or bird’s-eye-view images. These approaches benefit from efficient 2D convolutional networks and often achieve real-time inference. However, the projection process introduces geometric distortions and information loss, which can limit segmentation accuracy.

Voxel-based approaches~\cite{Cheng2021AF2S3NetAF,Zhu2021CylindricalAA,Hou2022PointtoVoxelKD} convert point clouds into structured 3D grids and process them using sparse convolutional networks~\cite{Tang2022TorchSparseEP,Choy20194DSC}. These methods have demonstrated strong performance by preserving spatial structure while enabling efficient computation. Nevertheless, discretization into voxels may lead to loss of fine geometric details when voxel resolution is reduced.

To leverage the strengths of different representations, recent works combine multiple modalities~\cite{Tang2020SearchingE3,Yan20222DPASS2P,Xu2021RPVNetAD,Ye2021DRINetAD}. For instance, SPVNAS~\cite{Tang2020SearchingE3} integrates point-based and voxel-based features within a unified architecture, while RPVNet~\cite{Xu2021RPVNetAD} employs a tri-branch network that jointly processes range-view, point-wise, and voxel features. These hybrid approaches improve segmentation accuracy by exploiting complementary geometric representations.

\subsection{Temporal LiDAR Reasoning}
To incorporate temporal context, several works attempt to aggregate information from multiple LiDAR frames. Early approaches~\cite{Schtt2022AbstractFF,Shi_2020_CVPR,Wang2022MetaRangeSegLS} integrate a small number of past frames to enhance segmentation accuracy. For example, SpSequenceNet~\cite{Shi_2020_CVPR} introduces cross-frame attention and interpolation mechanisms to combine features from consecutive scans, while MetaRangeSeg~\cite{Wang2022MetaRangeSegLS} incorporates residual depth information from previous frames. Although these methods improve short-term consistency, they operate within limited temporal windows and discard historical context beyond a few frames.

Other works explore recurrent formulations for temporal state maintenance. Duerr et al.~\cite{Duerr2020LiDARbasedR3} propose a recurrent model that aligns range-view features across streaming scans, while MemorySeg~\cite{Li2023MemorySegOL} introduces a persistent latent voxel memory for improved segmentation in long-range and occluded regions. StrObe~\cite{Frossard2020StrObeSO} similarly maintains a multi-scale BEV memory for incremental object detection. However, range-view and BEV representations are prone to projection artifacts and occlusion conflicts, and none of these approaches explicitly model recurring perceptual failures.

In contrast, DREAM augments LiDAR segmentation with persistent latent memory designed for long-horizon perception. Rather than storing raw observations or geometric maps, our framework maintains compact representations capturing both semantic experience and historical uncertainty. Retrieving relevant past contexts while suppressing recurring failure patterns enables robust, anticipatory perception, making DREAM one of the first approaches to handle LiDAR segmentation entirely within the latent space.
\section{Methodology}
Most LiDAR segmentation networks process each scan independently, ignoring temporal context critical for handling occlusions, sparsity, and recurring scene configurations. To address this, we introduce \textbf{DREAM} (Fig.~\ref{fig:dream_architecture}), a dual-memory reasoning framework for continual LiDAR semantic segmentation. DREAM treats segmentation as an experience-driven process where structured latent memories evolve, enabling anticipatory reasoning over streaming data in a strictly online regime.

\subsection{Problem Formulation}

\begin{figure*}[t]
    \centering
    \includegraphics[width=\textwidth]{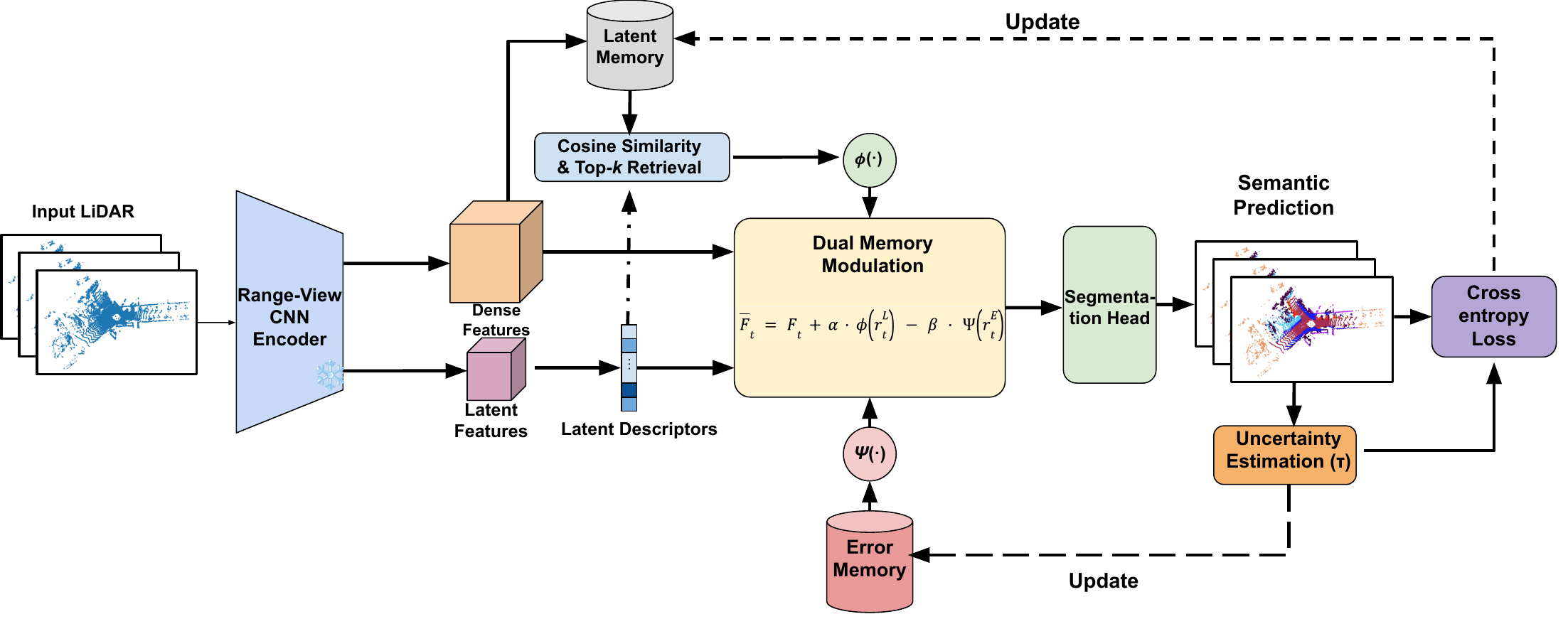}
    \caption{
    Overview of the proposed DREAM architecture. The network processes each LiDAR scan through hierarchical feature encoding and dual-memory reasoning modules that leverage past experience to guide current perception. 
    }
    \label{fig:dream_architecture}
\end{figure*}

Refer to Fig.~\ref{fig:dream_architecture} for the detailed architecture diagram of the proposed DREAM framework. We model segmentation as a recurrent reasoning process. 
At each timestep $t$, the model receives a LiDAR scan $X_t$ and produces semantic predictions $\hat{Y}_t$ conditioned on both the current input and an accumulated memory state $\mathcal{M}_{t-1}$. 
The system operates online: past scans are never revisited, and only compact latent representations are retained.

Given a temporally ordered sequence $\{X_t\}_{t=1}^{T}$, the model defines the mapping
\begin{equation}
\hat{Y}_t = f(X_t, \mathcal{M}_{t-1}; \theta),
\end{equation}
where $\theta$ denotes learnable parameters. 
After prediction, the memory is updated as
\begin{equation}
\mathcal{M}_t = \mathcal{U}(\mathcal{M}_{t-1}, X_t, \hat{Y}_t).
\end{equation}
The central challenge is to design a compact yet expressive representation for $\mathcal{M}_t$, along with mechanisms for retrieval and integration into perception.

\subsection{Perceptual Encoding}

Each raw LiDAR scan $X_t$ is projected into a range-view encoder 
$I_t \in \mathbb{R}^{5 \times H \times W}$ containing range, spatial coordinates $(x,y,z)$, and intensity. 
A convolutional backbone $g(\cdot)$ extracts hierarchical features:
\begin{equation}
F_t, Z_t = g(I_t),
\end{equation}
where $F_t \in \mathbb{R}^{C \times H' \times W'}$ are dense spatial features for segmentation and 
$Z_t \in \mathbb{R}^{D \times h \times w}$ are higher-level latent features. 
The backbone is frozen to stabilize representations and decouple long-term reasoning from low-level feature drift.

We summarize the latent representation via global average pooling:
\begin{equation}
\bar{z}_t = \text{GAP}(Z_t) \in \mathbb{R}^{D},
\end{equation}
which serves as the interface between perception and memory.

\subsection{Latent Memory Representation}
LiDAR observations are inherently partial, with objects often sparsely sampled, occluded, or incomplete across individual scans. Relying solely on the current frame leads to unstable predictions, particularly for distant or rarely observed objects. To address this, we introduce a latent perceptual memory that stores compressed semantic descriptors from past observations, enabling the model to retrieve relevant contextual information during inference.

Formally, we maintain a fixed-capacity latent memory bank
\begin{equation}
\mathcal{M}^L_t = \{ m^L_i \in \mathbb{R}^{D} \}_{i=1}^{K_L},
\end{equation}
where each entry encodes a compressed semantic abstraction of a previously observed scene. Given the global descriptor $\bar{z}_t$ extracted from the current scan, the system retrieves relevant past contexts through content-based addressing using cosine similarity:
\begin{equation}
s_i = \frac{\bar{z}_t^\top m^L_i}{\|\bar{z}_t\|\|m^L_i\|}.
\end{equation}
The top-$k$ most similar entries are selected to form the retrieved latent context
\begin{equation}
M^L_t \subset \mathcal{M}^L_{t-1}.
\end{equation}
By retrieving semantically related past descriptors, the model can incorporate previously observed contextual information when interpreting the current scan. This mechanism provides a compact form of long-term contextual memory that improves robustness to sparse observations, partial visibility, and recurring scene configurations.

\subsection{Error Memory Representation}

Complementary to perceptual memory, we introduce an error memory bank
\begin{equation}
\mathcal{M}^E_t = \{ m^E_j \in \mathbb{R}^{D} \}_{j=1}^{K_E},
\end{equation}
which explicitly encodes latent representations associated with unreliable predictions. 
Let $p_t(c|u)$ denote the predicted class probability at location $u$. 
Uncertain regions are identified using a confidence threshold $\tau$:
\begin{equation}
\mathcal{E}_t = \{ Z_t(u) \mid \max_c p_t(c|u) < \tau \}.
\end{equation}
Latent vectors from $\mathcal{E}_t$ are aggregated using mean pooling to produce a compact descriptor $\bar{e}_t \in \mathbb{R}^{D}$, which is then inserted into $\mathcal{M}^E_t$. 
While $\mathcal{M}^L_t$ captures stable semantic abstractions, $\mathcal{M}^E_t$ accumulates representations of recurring perceptual failures. 
By explicitly modeling these patterns, the system can anticipate similar uncertainty conditions in future observations and adjust its predictions accordingly.

\subsection{Dual Memory Modulation}

The retrieved latent memories $M^L_t$ and error memories $M^E_t$ are integrated into the dense feature space prior to classification. Error memory retrieval follows the same content-based addressing mechanism as the latent memory, where cosine similarity between the global descriptor $\bar{z}_t$ and stored error memory entries is computed and the most similar entry is selected using top-1 retrieval. Let $\phi(\cdot)$ and $\psi(\cdot)$ denote learnable projection functions mapping latent vectors to the dense feature space. We define the modulated features as
\begin{equation}
\tilde{F}_t = F_t + \alpha \cdot \phi(M^L_t) - \beta \cdot \psi(M^E_t),
\end{equation}
where $\alpha$ and $\beta$ control the strength of positive reinforcement and error suppression, respectively. 
In practice, $\alpha$ and $\beta$ are treated as fixed hyperparameters selected via validation to balance contextual enhancement against over-suppression. Empirically, moderate values ensure stable training and consistent gains across datasets. The additive formulation preserves spatial alignment while injecting global semantic priors derived from past experience. Crucially, modulation occurs before classification, allowing retrieved memories to influence perception before predictions are produced.

\subsection{Segmentation and Optimization}

The modulated features $\tilde{F}_t$ are fed to a lightweight segmentation head $h(\cdot)$:
\begin{equation}
\hat{Y}_t = h(\tilde{F}_t).
\end{equation}
Training minimizes a class-weighted cross-entropy loss
\begin{equation}
\mathcal{L}_t = \text{CE}(\hat{Y}_t, Y_t),
\end{equation}
where only the segmentation head and modulation layers are optimized, while the backbone remains frozen.

\subsection{Online Memory Update}

After prediction, the global descriptor $\bar{z}_t$ is inserted into $\mathcal{M}^L_t$, and the aggregated error descriptor $\bar{e}_t$ derived from uncertain latents $\mathcal{E}_t$ is inserted into $\mathcal{M}^E_t$. 
Both memories operate with fixed capacity using a replacement strategy (e.g., FIFO), ensuring bounded memory growth while retaining long-term experience in compressed form.

\subsection{Memory-Conditioned Perception}

DREAM formulates LiDAR semantic segmentation as a perception problem conditioned on accumulated context from streaming observations. Rather than processing each scan independently, the model maintains two latent memory banks, $\mathcal{M}^L_t$ and $\mathcal{M}^E_t$, which store compact representations of previously observed semantic structures and historical low-confidence predictions. At each timestep, a global latent descriptor retrieves relevant entries through content-based addressing, and the retrieved memories modulate the dense feature representation prior to classification. This mechanism allows the model to incorporate contextual information from past observations while maintaining a feed-forward inference pipeline. As a result, DREAM integrates long-horizon contextual cues into LiDAR perception in a computationally efficient manner, enabling temporally aware semantic segmentation in streaming autonomous driving scenarios.

\section{Experimental Setup}

We evaluate the proposed framework on three large-scale autonomous driving datasets: SemanticKITTI~\cite{behley2019iccv}, nuScenes~\cite{Caesar2019nuScenesAM}, and PandaSet~\cite{Xiao2021PandaSetAS}. These datasets are collected using different LiDAR sensors and capture diverse urban environments, making them suitable for evaluating perception systems operating in streaming conditions. Our experiments focus on assessing the effectiveness of persistent memory for long-horizon perception and its ability to improve robustness under sparse observations, occlusions, and dynamic scene changes. In addition to comparisons with existing methods, we perform ablation studies to analyze the contribution of each component of the proposed dual-memory design.

\subsection{Datasets}
We evaluate on three large-scale autonomous driving datasets. \textbf{SemanticKITTI}~\cite{behley2019iccv} comprises 22 sequences captured with a Velodyne HDL-64E in Germany, with sequences 0--10 for training and 11--21 for testing. Labels are mapped to 19 classes (single-scan) and 25 classes (multi-scan), where we primarily report multi-scan results as our method models temporal context. \textbf{nuScenes}~\cite{Caesar2019nuScenesAM} contains 1000 driving scenes from Boston and Singapore (700/150/150 train/val/test split), captured with a Velodyne HDL-32E at 2\,Hz, with 31 classes mapped to 16 categories. \textbf{PandaSet}~\cite{Xiao2021PandaSetAS} provides temporally continuous sequences recorded in Silicon Valley using a Hesai Pandar64, with 103 sequences and fine-grained classes grouped into 14 categories following~\cite{Duerr2020LiDARbasedR3}.

\subsection{Evaluation Protocol}

During evaluation, each sequence is processed in temporal order from the first frame to the last. The memory state is updated online during inference, allowing the model to accumulate experience across the sequence. Performance is evaluated using Intersection-over-Union (IoU) \cite{Rezatofighi2019GeneralizedIO}, with mean IoU (mIoU) across all semantic classes reported as the primary metric. In addition, we report seen-class mIoU, which averages IoU only over classes present in the evaluated sequence, as well as class-wise IoU scores for detailed analysis.

\subsection{Implementation Details}
Training proceeds in two stages: the segmentation head is first trained for 50 epochs in a single-scan setting, after which the dual-memory mechanism and modulation layers are fine-tuned for 20 additional epochs in an online recurrent manner. The backbone remains frozen throughout. Each LiDAR scan is projected into a $64 \times 1024$ range-view with five channels (range, $x$, $y$, $z$, intensity). A pre-trained ResNet-18 extracts dense features from the third residual stage and a 512-dimensional global descriptor from the final stage for memory retrieval. The latent and error memory banks store $K_L{=}64$ and $K_E{=}32$ vectors ($D{=}512$) respectively, using cosine similarity for top-1 retrieval and FIFO replacement. The error memory is updated via low-confidence regions. Optimization uses Adam ($\text{lr}=5\times10^{-4}$) with class-weighted cross-entropy and mixed precision training. Augmentations include global scaling, random translation, and vertical-axis rotation. The model is trained on 8 NVIDIA A100 GPUs (80GB) using distributed data parallelism.

\begin{table*}[t]
\centering
\renewcommand{\arraystretch}{1.1}
\resizebox{\textwidth}{!}{
\begin{tabular}{lc*{25}{c}}
\toprule
\textbf{Method} & \textbf{mIoU}
& \rotatebox{90}{Car}
& \rotatebox{90}{Bicycle}
& \rotatebox{90}{Motorcycle}
& \rotatebox{90}{Truck}
& \rotatebox{90}{Other Vehicle}
& \rotatebox{90}{Person}
& \rotatebox{90}{Bicyclist}
& \rotatebox{90}{Motorcyclist}
& \rotatebox{90}{Road}
& \rotatebox{90}{Parking}
& \rotatebox{90}{Sidewalk}
& \rotatebox{90}{Other Ground}
& \rotatebox{90}{Building}
& \rotatebox{90}{Fence}
& \rotatebox{90}{Vegetation}
& \rotatebox{90}{Trunk}
& \rotatebox{90}{Terrain}
& \rotatebox{90}{Pole}
& \rotatebox{90}{Traffic Sign}
& \rotatebox{90}{Car (m)}
& \rotatebox{90}{Bicyclist (m)}
& \rotatebox{90}{Person (m)}
& \rotatebox{90}{Motorcyclist (m)}
& \rotatebox{90}{Other Vehicle (m)}
& \rotatebox{90}{Truck (m)} \\
\midrule
TangentConv~\cite{Tatarchenko2018TangentCF}
& 34.1 & 84.9 & 2.0 & 18.2 & 21.1 & 18.5 & 1.6 & 0.0 & 0.0 & 83.9 & 38.3 & 64.0 & 15.3 & 85.8 & 49.1 & 79.5 & 43.2 & 56.7 & 36.4 & 31.2 & 40.3 & 1.1 & 6.4 & 1.9 & 30.1 & 42.2 \\
DarkNet53Seg~\cite{Behley2019ADF}
& 41.6 & 84.1 & 30.4 & 32.9 & 20.2 & 20.7 & 7.5 & 0.0 & 0.0 & 91.6 & 64.9 & 75.3 & 27.5 & 85.2 & 56.5 & 78.4 & 50.7 & 64.8 & 38.1 & 53.3 & 61.5 & 14.1 & 15.2 & 0.2 & 28.9 & 37.8 \\
KPConv~\cite{Thomas2019KPConvFA}
& 51.2 & 93.7 & 44.9 & 47.2 & 42.5 & 38.6 & 21.6 & 0.0 & 0.0 & 86.5 & 58.4 & 70.5 & 26.7 & 90.8 & 64.5 & 84.6 & 70.3 & 66.0 & 57.0 & 53.9 & 69.4 & 67.4 & 67.5 & 47.2 & 4.7 & 5.8 \\
Cylinder3D~\cite{Zhu2021CylindricalAA}
& 52.5 & 94.6 & 67.6 & 63.8 & 41.3 & 38.8 & 12.5 & 1.7 & 0.2 & 90.7 & \textbf{65.0} & 74.5 & \textbf{32.3} & 92.6 & 66.0 & \textbf{85.8} & 72.0 & 68.9 & 63.1 & 61.4 & 74.9 & 68.3 & 65.7 & 11.9 & 0.1 & 0.0 \\
\midrule
SpSequenceNet~\cite{Shi_2020_CVPR}
& 43.1 & 88.5 & 24.0 & 26.2 & 29.2 & 22.7 & 6.3 & 0.0 & 0.0 & 90.1 & 57.6 & 73.9 & 27.1 & 91.2 & 66.8 & 84.0 & 66.0 & 65.7 & 50.8 & 48.7 & 53.2 & 41.2 & 26.2 & 36.2 & 2.3 & 0.1 \\
TemporalLidarSeg~\cite{Duerr2020LiDARbasedR3}
& 47.0 & 92.1 & 47.7 & 40.9 & 39.2 & 35.0 & 14.4 & 0.0 & 0.0 & 91.8 & 59.6 & \textbf{75.8} & 23.2 & 89.8 & 63.8 & 82.3 & 62.5 & 64.7 & 52.6 & 60.4 & 68.2 & 42.8 & 40.4 & 12.9 & 12.4 & 2.1 \\
TemporalLatticeNet~\cite{TemporalLatticeNet}
& 47.1 & 91.6 & 35.4 & 36.1 & 26.9 & 23.0 & 9.4 & 0.0 & 0.0 & 91.5 & 59.3 & 75.3 & 27.5 & 89.6 & 65.3 & 84.6 & 66.7 & 70.4 & 57.2 & 60.4 & 59.7 & 41.7 & 51.0 & 48.8 & 5.9 & 0.0 \\
Meta-RangeSeg~\cite{Wang2022MetaRangeSegLS}
& 49.7 & 90.8 & 50.0 & 49.5 & 29.5 & 34.8 & 16.6 & 0.0 & 0.0 & 90.8 & 62.9 & 74.8 & 26.5 & 89.8 & 62.1 & 82.8 & 65.7 & 66.5 & 56.2 & 64.5 & 69.0 & 60.4 & 57.9 & 22.0 & 16.6 & 2.6 \\
MemorySeg~\cite{Li2023MemorySegOL}
& 58.3 & 94.0 & 68.3 & 68.8 & 51.3 & 40.9 & 27.0 & 0.3 & 2.8 & 89.9 & 64.3 & 74.8 & 29.2 & 92.2 & 69.3 & 84.8 & 75.1 & 70.1 & 65.5 & \textbf{68.5} & 71.7 & \textbf{74.4} & \textbf{71.7} & 73.9 & 15.1 & 13.6 \\
\midrule
\textbf{DREAM (Ours)}
& \textbf{61.7} & \textbf{97.6} & \textbf{69.1} & \textbf{68.9} & \textbf{55.7} & \textbf{43.1} & \textbf{29.0} & \textbf{1.7} & \textbf{3.2} & \textbf{93.5} & 62.6 & 74.8 & 29.2 & 92.2 & 69.3 & 84.8 & \textbf{77.3} & \textbf{71.0} & \textbf{65.6} & 68.4 & 71.5 & 70.3 & 71.5 & \textbf{74.3} & \textbf{37.6} & \textbf{49.7} \\
\bottomrule
\end{tabular}
}
\caption{Comparison with state-of-the-art LiDAR semantic segmentation methods on the \textbf{SemanticKITTI} multi-scan test benchmark. (m) denotes moving classes. Metrics are reported in mIoU (\%).}
\label{tab:sota_semantickitti}
\end{table*}

\begin{table*}[t]
\centering
\small
\setlength{\tabcolsep}{4pt}
\resizebox{\textwidth}{!}{
\begin{tabular}{lcc|cccccccccccccccc}
\toprule
\textbf{Method} & \textbf{mIoU} & \textbf{FW-mIoU} &
\rotatebox{90}{Barrier} &
\rotatebox{90}{Bicycle} &
\rotatebox{90}{Bus} &
\rotatebox{90}{Car} &
\rotatebox{90}{Construction} &
\rotatebox{90}{Motorcycle} &
\rotatebox{90}{Pedestrian} &
\rotatebox{90}{Traffic Cone} &
\rotatebox{90}{Trailer} &
\rotatebox{90}{Truck} &
\rotatebox{90}{Drivable} &
\rotatebox{90}{Other Flat} &
\rotatebox{90}{Sidewalk} &
\rotatebox{90}{Terrain} &
\rotatebox{90}{Manmade} &
\rotatebox{90}{Vegetation} \\
\midrule

PolarNet~\cite{Zhang2020PolarNetAI} 
& 69.4 & 87.4 
& 72.2 & 16.8 & 77.0 & 86.5 & 51.1 & 69.7 & 64.8 & 54.1 & 69.7 & 63.5 & 96.6 & 67.1 & 77.7 & 72.1 & 87.1 & 84.5 \\

Cylinder3D~\cite{Zhu2021CylindricalAA} 
& 77.2 & 89.9 
& 82.8 & 29.8 & 84.3 & 89.4 & 63.0 & \textbf{79.3} & 77.2 & 73.4 & 84.6 & 69.1 & 97.7 & \textbf{70.2} & 80.3 & 75.5 & 90.4 & 87.6 \\

SPVCNN~\cite{Tang2020SearchingE3} 
& 77.4 & 89.7 
& 80.0 & 30.0 & 91.9 & 90.8 & 64.7 & 79.0 & 75.6 & 70.9 & 81.0 & 74.6 & 97.4 & 69.2 & 80.0 & 76.1 & 89.3 & 87.1 \\

(AF)$^2$-S3Net~\cite{Cheng2021AF2S3NetAF} 
& 78.3 & 88.5 
& 78.9 & \textbf{52.2} & 89.9 & 84.2 & \textbf{77.4} & 74.3 & 77.3 & 72.0 & 83.9 & 73.8 & 97.1 & 66.5 & 77.5 & 74.0 & 87.7 & 86.8 \\

MemorySeg~\cite{Li2023MemorySegOL} 
& 80.6 & 91.4 
& 84.9 & 40.2 & 91.2 & 92.4 & 71.2 & 73.5 & \textbf{85.9} & 77.8 & 88.0 & 76.4 & \textbf{97.9} & 69.0 & 81.2 & 77.6 & 92.6 & 89.7 \\

\midrule
\textbf{DREAM (Ours)} 
& \textbf{81.8} & \textbf{91.4} 
& \textbf{86.3} & 40.2 & \textbf{92.3} & \textbf{94.5} & \textbf{79.4} & 73.9 & 84.7 & \textbf{78.0} & \textbf{88.4} & \textbf{79.1} & 96.2 & 68.3 & \textbf{81.9} & 76.3 & \textbf{93.5} & \textbf{90.4} \\

\bottomrule
\end{tabular}
}
\caption{Comparison with state-of-the-art LiDAR-only semantic segmentation methods on the \textbf{nuScenes} test set~\cite{Caesar2019nuScenesAM}. Metrics are reported in mIoU (\%). Best results are highlighted in \textbf{bold}.}
\label{tab:sota_nuscenes}
\end{table*}

\begin{table*}[t]
\centering
\small
\setlength{\tabcolsep}{4pt}
\resizebox{\textwidth}{!}{
\begin{tabular}{l c ccccccccccccccc}
\toprule
\textbf{Method} & \textbf{mIoU}
& \rotatebox{90}{Car}
& \rotatebox{90}{Bicycle}
& \rotatebox{90}{Motorcycle}
& \rotatebox{90}{Truck}
& \rotatebox{90}{Other Vehicle}
& \rotatebox{90}{Person}
& \rotatebox{90}{Road}
& \rotatebox{90}{Road Barriers}
& \rotatebox{90}{Sidewalk}
& \rotatebox{90}{Building}
& \rotatebox{90}{Vegetation}
& \rotatebox{90}{Terrain}
& \rotatebox{90}{Background}
& \rotatebox{90}{Traffic Sign} \\
\midrule

SqueezeSegv3~\cite{Xu2020SqueezeSegV3SC} & 55.7
& 92.8 & 24.1 & 18.0 & 36.5 & 54.3 & 63.0 & 91.1 & 11.9 & 71.3 & 86.2 & 85.0 & 61.3 & 63.2 & 20.6 \\

SalsaNext~\cite{Cortinhal2020SalsaNextFS} & 57.8
& 92.1 & 40.7 & 31.7 & 28.7 & 56.2 & 69.0 & 90.0 & 22.6 & 67.1 & 85.6 & 83.4 & 58.5 & 63.3 & 20.6 \\

TemporalLidarSeg~\cite{Duerr2020LiDARbasedR3} & 60.0
& 93.7 & 33.6 & 38.0 & 37.1 & 59.9 & 72.0 & 91.1 & 14.6 & 70.6 & 88.2 & 88.4 & 63.8 & 68.4 & 20.7 \\

SPVCNN~\cite{Tang2020SearchingE3} & 64.7
& 95.8 & 38.1 & 46.3 & 44.0 & 74.1 & 78.6 & 91.2 & \textbf{28.3} & 70.3 & 87.2 & 87.5 & 61.5 & 67.5 & 35.6 \\

\textbf{MemorySeg}~\cite{Li2023MemorySegOL} & 70.3
& 97.2 & 60.2 & 58.4 & 62.9 & 74.3 & 82.6 & 92.1 & 27.7 & 74.1 & 89.4 & 90.7 & \textbf{64.9} & 72.8 & 36.4 \\

\midrule
\textbf{DREAM (Ours)} & \textbf{71.0}
& \textbf{97.8} & \textbf{60.5} & \textbf{59.1} & \textbf{64.0} & \textbf{75.4} & \textbf{82.9} & \textbf{93.6} & 28.1 & \textbf{75.4} & \textbf{90.7} & \textbf{89.2} & 64.5 & \textbf{74.6} & \textbf{37.1} \\

\bottomrule
\end{tabular}
}
\caption{Comparison with state-of-the-art LiDAR semantic segmentation methods on the PandaSet test set~\cite{Xiao2021PandaSetAS}. Metrics are reported in mIoU (\%). Best results are highlighted in \textbf{bold}.}
\label{tab:sota_pandaset}
\end{table*}

\subsection{Quantitative Analysis}
We evaluate DREAM on SemanticKITTI, nuScenes, and PandaSet, with results summarized in Tables~\ref{tab:sota_semantickitti}, \ref{tab:sota_nuscenes}, and \ref{tab:sota_pandaset}.

On SemanticKITTI (Tab.~\ref{tab:sota_semantickitti}), DREAM achieves 61.7\% mIoU, outperforming MemorySeg (58.3\%) and Meta-RangeSeg (49.7\%). Gains are most pronounced in dynamic categories (moving cars, bicyclists, pedestrians, motorcyclists) and structural classes (building, trunk, terrain), reflecting the benefit of temporal reasoning via the dual-memory design.

On nuScenes (Tab.~\ref{tab:sota_nuscenes}), DREAM achieves \textbf{81.8\%} mIoU, with notable improvements in barrier, pedestrian, and traffic cone categories, suggesting the memory mechanism effectively addresses sparsity and scale challenges.

On PandaSet (Tab.~\ref{tab:sota_pandaset}), DREAM achieves \textbf{71.0\%} mIoU, performing consistently across foreground classes (truck, person, traffic sign) and background classes (sidewalk, terrain), confirming generalization across sensor configurations and urban environments.

Across all benchmarks, persistent latent memory enables robust segmentation under sparse observations, occlusions, and dynamic scene changes.

\subsection{Comparison Against State-of-the-Art}
We compare DREAM with prior LiDAR-only methods across all three benchmarks (Tab.~\ref{tab:sota_semantickitti}--\ref{tab:sota_pandaset}).

On SemanticKITTI (Tab.~\ref{tab:sota_semantickitti}), DREAM outperforms Cylinder3D (52.5\%), Meta-RangeSeg (49.7\%), and MemorySeg (58.3\%), with the largest gains in dynamic classes (moving cars, bicyclists, pedestrians, motorcyclists), demonstrating effective temporal modeling via dual memory.

On nuScenes (Tab.~\ref{tab:sota_nuscenes}), DREAM sets a new state-of-the-art among LiDAR-only methods, surpassing MemorySeg (80.6\%) and (AF)$^2$-S3Net (78.3\%), with notable improvements in small-object categories such as barrier, pedestrian, and traffic cone.

On PandaSet (Tab.~\ref{tab:sota_pandaset}), DREAM achieves the best overall performance over MemorySeg (70.3\%) and SPVCNN (64.7\%), with consistent gains across foreground and background categories including truck, person, and sidewalk.

Across all datasets, persistent latent memory reasoning consistently outperforms both single-frame and short-term temporal fusion approaches.

\begin{figure*}[t]
    \centering
    \includegraphics[width=\linewidth]{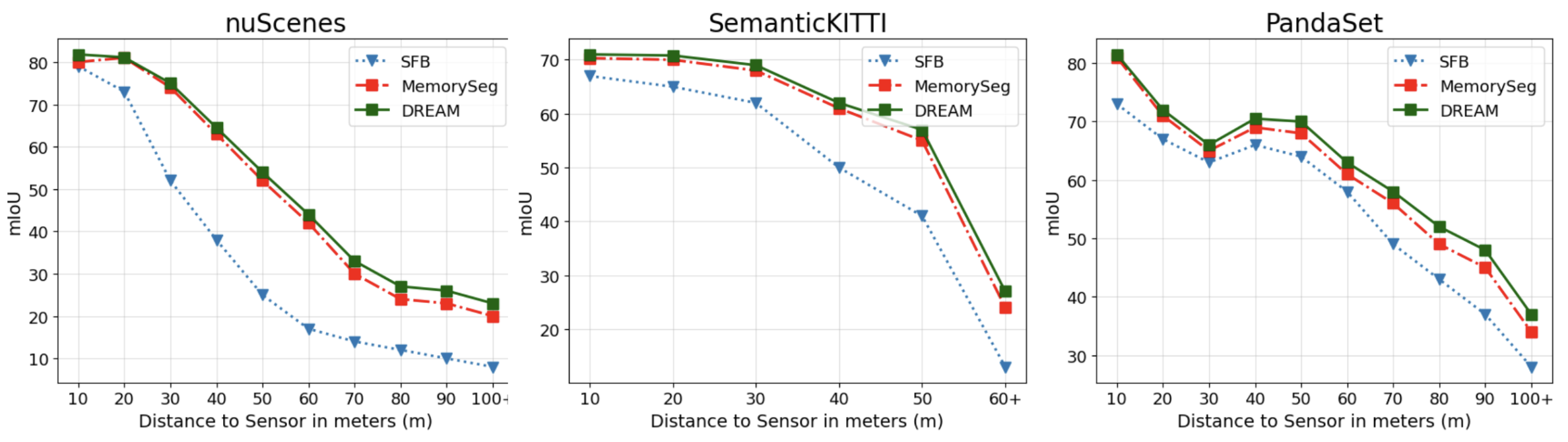}
    \caption{Distance-based performance comparison of SFB (dooted blue), MemorySeg (solid red), and DREAM (solid green) across nuScenes, SemanticKITTI, and PandaSet. DREAM demonstrates consistently improved robustness at mid- and long-range distances, with gains increasing as point sparsity and occlusion become more severe.}
    \label{fig:distance_analysis}
\end{figure*}

\subsection{Qualitative Analysis}

\subsubsection{Temporal Consistency Analysis}

\begin{figure}[!htbp]
\centering
\includegraphics[width=\linewidth]{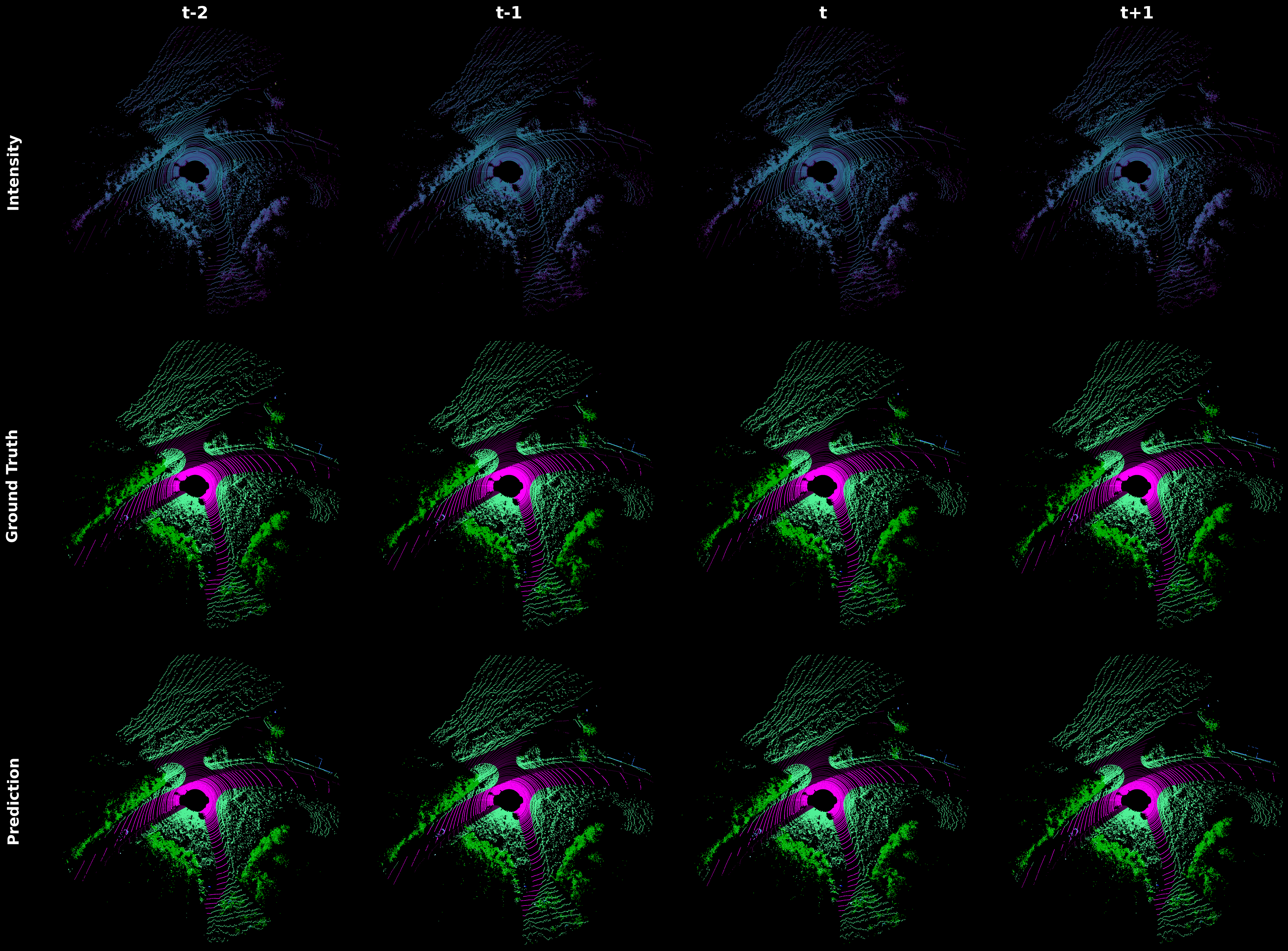}
\caption{
Temporal qualitative results across consecutive LiDAR frames.Columns correspond to four consecutive frames ($t-2$, $t-1$, $t$, $t+1$). The top row shows the input LiDAR intensity visualization, the middle row shows the ground-truth semantic labels, and the bottom row shows the predictions produced by the proposed DREAM framework.
}
\label{fig:temporal_sequence}
\end{figure}

Fig.~\ref{fig:temporal_sequence} presents qualitative results across four consecutive LiDAR frames ($t-2$, $t-1$, $t$, $t+1$). The top row shows the input LiDAR intensity visualization, while the middle and bottom rows illustrate the ground-truth semantic labels and predictions from DREAM, respectively.

Despite the sparse and partially observed nature of individual LiDAR scans, the predicted semantic structure remains stable across time. In particular, large structural classes such as road and vegetation are consistently segmented across frames. This temporal stability demonstrates the benefit of the proposed memory-based reasoning mechanism, which allows the model to leverage historical context when interpreting the current observation.

\subsubsection{Occlusion Recovery}

\begin{figure}[!]
\centering
\includegraphics[width=\linewidth]{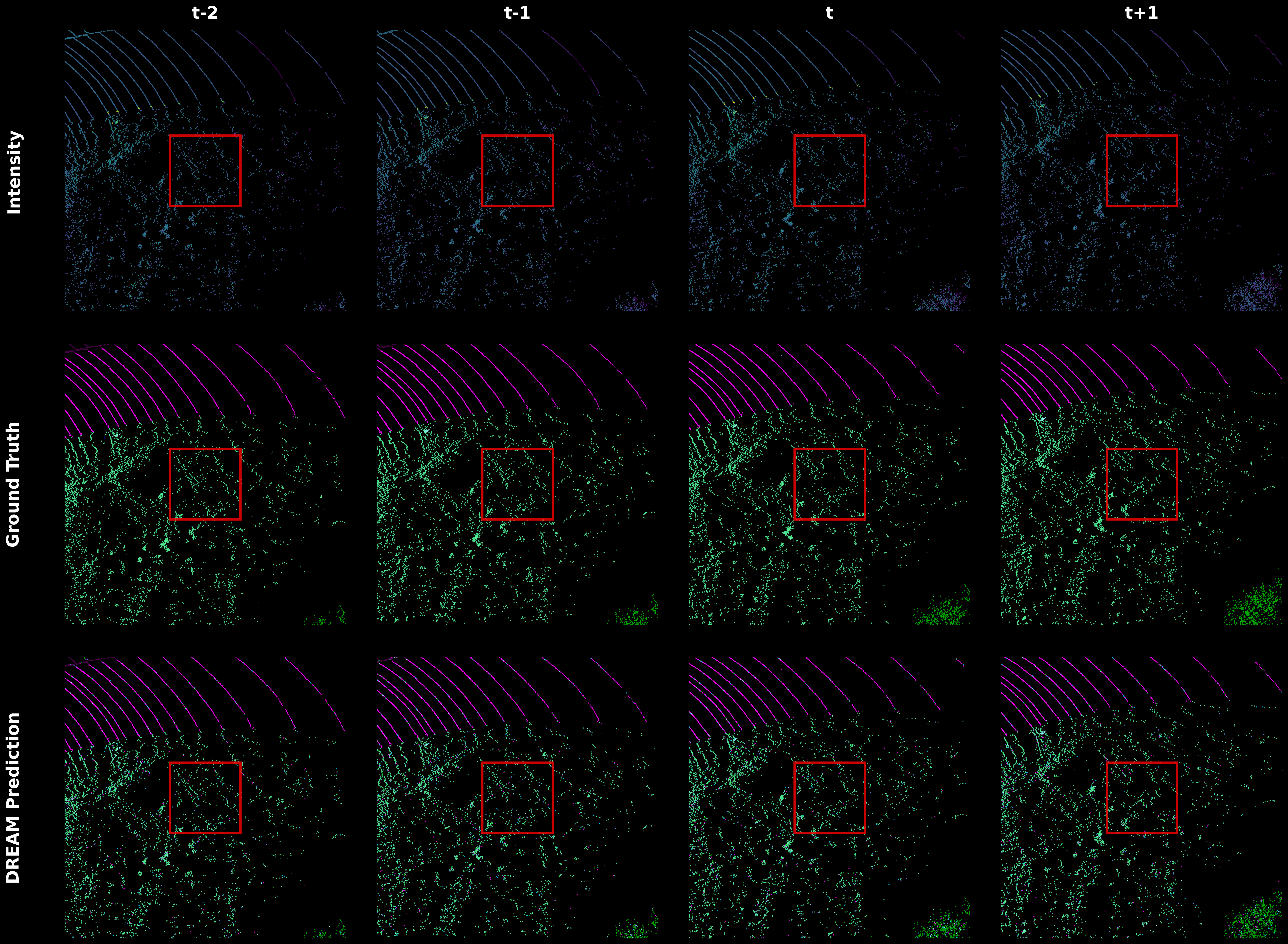}
\caption{
Qualitative example illustrating occlusion recovery across consecutive frames. As objects become partially occluded, DREAM maintains stable semantic predictions by leveraging temporal context stored in latent memory. }
\label{fig:occlusion_recovery}
\end{figure}

Fig.~\ref{fig:occlusion_recovery} illustrates the behavior of the DREAM framework under partial occlusions. The top row shows the input LiDAR intensity visualization, where the highlighted region exhibits varying point density across consecutive frames due to occlusion and sparse observations.

The middle row shows the ground-truth semantic labels, while the bottom row presents predictions produced by DREAM. Despite incomplete observations in individual LiDAR scans, DREAM maintains temporally consistent semantic predictions with 61.7\% mIoU and 81.8\% mIoU on SemanticKITTI and nuScenes datasets, respectively, across frames, demonstrating the proposed memory mechanism's ability to recover semantic structure in occluded or sparsely observed regions.

\begin{table}[!]
\centering
\small
\setlength{\tabcolsep}{8pt}
\begin{tabular}{lcc}
\toprule
\textbf{Method} & \textbf{Params (M)} & \textbf{FPS} \\
\midrule
SFB & 14.2 & 29 \\
DREAM (Full) & 16.8 & 25 \\
\bottomrule
\end{tabular}
\caption{
Parameter count and inference speed comparison on SemanticKITTI.
DREAM introduces a modest increase in the number of parameters and runtime while maintaining real-time performance.
}
\label{tab:runtime}
\end{table}

\subsection{Ablation Studies}

\subsubsection{Importance of the Memory}

To evaluate the impact of persistent memory, we compare DREAM against a single-frame baseline (SFB) that removes both latent perceptual memory and error memory components. The comparison is performed using the distance-based evaluation shown in Fig.~\ref{fig:distance_analysis}. Results indicate that incorporating memory significantly improves segmentation performance, particularly at larger distances where LiDAR measurements are sparse and noisy. This demonstrates that maintaining a persistent latent representation of past observations helps the model recover semantic context that may be missing in individual frames.


\subsubsection{Distance-based Analysis}
Fig.~\ref{fig:distance_analysis} highlights the benefit of persistent dual memory at longer ranges, where LiDAR point clouds become increasingly sparse. We compare DREAM against a single-frame baseline (SFB) that removes both memory modules.

DREAM consistently outperforms the SFB across all distance bins, with the most pronounced gains at 80m: 26.47\% mIoU on nuScenes and 52.33\% mIoU on PandaSet, where point density is lowest. The progressive improvement with distance confirms that persistent memory becomes more beneficial as perceptual difficulty increases, enhancing semantic consistency through compact scene representations and explicit uncertainty modeling.

\begin{table}[!]
\centering
\small
\setlength{\tabcolsep}{6pt}
\begin{tabular}{lcc}
\toprule
\textbf{Method} & \textbf{mIoU (\%)} \\
\midrule
SFB & 58.4 \\
DREAM w/o Error Memory & 60.9 \\
DREAM (Full) & \textbf{61.5} \\
\bottomrule
\end{tabular}
\caption{
Ablation study on the SemanticKITTI multi-scan validation set.
Latent perceptual memory accounts for the majority of the improvement over the single-frame baseline (SFB).
The proposed error memory further enhances performance, particularly in long-range regions (60m+), 
where point sparsity and occlusions are more severe.
}
\label{tab:error_memory_ablation}
\end{table}

\subsubsection{Effect of Error Memory}
\label{sec:error_memory}

We evaluate the contribution of the proposed error memory on the SemanticKITTI multi-scan validation set (Tab.~\ref{tab:error_memory_ablation}). Compared to the single-frame baseline (SFB), introducing latent perceptual memory improves mIoU from 78.6\% to 81.0\%, highlighting the benefit of persistent semantic aggregation. Adding the error memory further increases performance to 81.8\%, with more pronounced gains in long-range regions, where mIoU improves from 24.8\% to 27.3\%. These results indicate that explicitly modeling recurrent low-confidence patterns complements spatial memory and enhances robustness under sparse and occluded observations.

\subsubsection{Parameter Count.} DREAM introduces only a modest computational overhead compared to the single-frame baseline (SFB). 
As shown in Tab.~\ref{tab:runtime}, the proposed dual-memory design increases the parameter count from 14.2M to 16.8M (approximately 18\%) and reduces inference speed from 29 FPS to 25 FPS (approximately 15\%), while maintaining real-time performance.

\section{Conclusion}
We presented DREAM, a dual-memory reasoning framework for continual LiDAR semantic segmentation that reformulates perception as a temporally grounded process. DREAM maintains two complementary memory banks: a perceptual memory that accumulates stable semantic abstractions and an error memory that records recurring failure patterns. Retrieved memories modulate dense features before classification, reinforcing consistent semantics while suppressing previously observed errors. The framework operates strictly online, avoids scan replay, and keeps the backbone frozen, yielding a bounded-memory design suitable for real-world deployment. Experiments on SemanticKITTI, nuScenes, and PandaSet demonstrate state-of-the-art performance, with the most significant gains in dynamic and small-object categories. DREAM advances LiDAR segmentation from stateless prediction to memory-augmented reasoning, offering a scalable foundation for long-horizon autonomous perception.
{
    \small
    \bibliographystyle{ieeenat_fullname}
    \bibliography{main}
}


\end{document}